# Recursive InPainting (RIP): how much information is lost under recursive inferences?

Javier Conde[1] · Miguel Gonzalez[1] · Gonzalo Martínez[2] · Fernando Moral[3] · Elena Merino-Gomez[4] · Pedro Reviriego[1]

## Abstract
The rapid adoption of generative artificial intelligence (AI) is accelerating content creation and modification. For example, variations of a given content, be it text or images, can be created almost instantly and at a low cost. This will soon lead to the majority of text and images being created directly by AI models or by humans assisted by AI. This poses new risks; for example, AI-generated content may be used to train newer AI models and degrade their performance, or information may be lost in the transformations made by AI which could occur when the same content is processed over and over again by AI tools. An example of AI image modifications is inpainting in which an AI model completes missing fragments of an image. The incorporation of inpainting tools into photo editing programs promotes their adoption and encourages their recursive use to modify images. Inpainting can be applied recursively, starting from an image, removing some parts, applying inpainting to reconstruct the image, revising it, and then starting the inpainting process again on the reconstructed image, etc. This paper presents an empirical evaluation of recursive inpainting when using one of the most widely used image models: Stable Diffusion. The inpainting process is applied by randomly selecting a fragment of the image, reconstructing it, selecting another fragment, and repeating the process a predefined number of iterations. The images used in the experiments are taken from a publicly available art data set and correspond to different styles and historical periods. Additionally, photographs are also evaluated as a reference. The modified images are compared with the original ones by both using quantitative metrics and performing a qualitative analysis. The results show that recursive inpainting in some cases modifies the image so that it still resembles the original one while in others leads to image degeneration, so ending with a non-meaningful image. The outcome of the recursive inpainting process depends on several factors, such as the type of image, the size of the inpainting masks, and the number of iterations. The results of our evaluation illustrate how information can be lost due to successive AI transformations. The evaluation of additional models, images, and inpainting sequences is needed to confirm whether this observation is generally applicable or if it occurs only in some models and settings.

✉ Gonzalo Martínez
gonzmart@pa.uc3m.es

✉ Pedro Reviriego
pedro.reviriego@upm.es

Javier Conde
javier.conde.diaz@upm.es

Miguel Gonzalez
miguel.gonsaiz@upm.es

Fernando Moral
fmoral@nebrija.es

Elena Merino-Gomez
elena.merino.gomez@uva.es

[1] Technical University of Madrid, Madrid, Spain
[2] Carlos III University of Madrid, Madrid, Spain
[3] Nebrija University, Madrid, Spain
[4] University of Valladolid, Valladolid, Spain

Springer

# 1 Introduction

Generative Artificial Intelligence (AI) has taken center stage in the last two years and triggered a new technology revolution. Generative AI models can generate text, audio, images, or video and can be used in many transformative applications. Among the AI tools, Large Language Models (LLMs) such as GPT4 (Achiam et al. 2023), which can answer questions, summarize, translate, and paraphrase texts, and text-to-image generators such as DALL-E (Ramesh et al. 2021), which can create images for almost any text description have attracted the interest of the public with hundreds of millions of users. These tools are now used, for example, to help in storytelling creation (Antony and Huang 2024). As our decisions depend more and more on AI systems (Cau et al. 2023), it is important to make sure that they provide trustworthy information.

These tools have achieved unprecedented performance levels in most tasks, and evaluating their performance is a key issue. In the case of LLMs, many benchmarks have been proposed to assess their knowledge of different topics, their ability to solve math (Cobbe et al. 2021) or reasoning problems (Srivastava et al. 2022), or their language understanding (Guo et al. 2023). Those benchmarks are used to compare models, and when a new model is introduced, typically performance over the most common benchmarks is reported (Touvron et al. 2023). In the case of image generation tools, a number of metrics have been proposed to evaluate performance, such as the Fréchet Inception Distance (FID) (Heusel et al. 2017), precision and recall (Kynkäänniemi et al. 2019) or diversity and coverage (Naeem et al. 2020) that try to capture how close are generated images to real ones and how well generated-images cover the range of real images. Another feature supported by some AI image generation tools and implemented with ad-hoc AI models is inpainting (Yang et al. 2023). In this case, the AI tool is given an image with missing fragments and has to fill them to complete the image, or to regenerate fragments of an image to produce a new version according to the prompt of the user. Current Generative AI models are able to generate images from scratch based on user prompts, but it is difficult for them to meet all user requirements. In this regard, inpainting systems can assist in post-processing the image and making modifications to it. Photo editing programs like Photoshop already include inpainting tools based on AI models (Batley et al. 2024).

Evaluating the quality of the content generated by AI is not only relevant to compare AI models or to assess their progress in different tasks. The widespread adoption of generative AI is transforming the nature of content on the Internet. Texts and images generated by AI are now widespread and, in some cases, dominant, and the trend is expected to continue in the next years. As the cost and time needed to transform content decreases, we can expect that the variations and transformations of text and images become more frequent and are applied recursively. For example, some news generated by a source will be translated and processed into other languages and those may be rephrased; all this is done by AI systems in a negligible amount of time. This easiness in content generation may in some cases lead to loss of information which could eventually cause misinformation if content is not properly checked. This has implications for AI models, as we now have to understand how they perform when they process the same information recursively. This has also implications for newer AI models as they are commonly trained on data that is scraped from the Internet, so a loop is created where newer AI models are trained with data generated with previous AI models (Martínez et al. 2023). This can lead to worse performance or even the collapse of AI models (Dohmatob et al. 2024) and has triggered research on the stability of AI models when trained with their own data (Bertrand et al. 2023; Briesch et al. 2023).

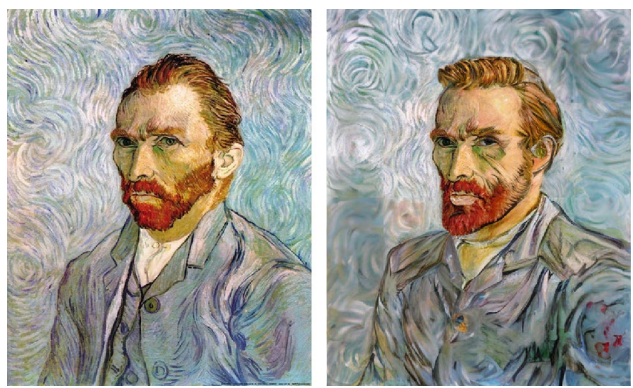

**Fig. 1** Example of the Recursive InPainting (RIP) process on "Van Gogh, Self portrait, 1889": original and outcome versions side by side

There are other scenarios that can lead to information loss, for example, when the input to the AI model is, for instance, an image and the output is also an image, as in the case of inpainting, the AI model can be used recursively on its output, creating a loop, an example of this process[1] is shown in Fig. 1. This models successive transformations of content by AI models and in this case, there is no training, only inferences that are applied recursively. Applying recursive modifications to an image can cause it to lose its original characteristics, leading to versions that are more prevalent in the model's training dataset or influenced by the model's inherent biases. Analyzing the impact of these recursive calls to the AI model on the generated content is of interest to understand whether the AI models are stable or also lead to degeneration and information loss.

---

[1] The recursive inpainting process can be tested in a publicly available demo at https://huggingface.co/spaces/GING-UPM/recursive_inpainting

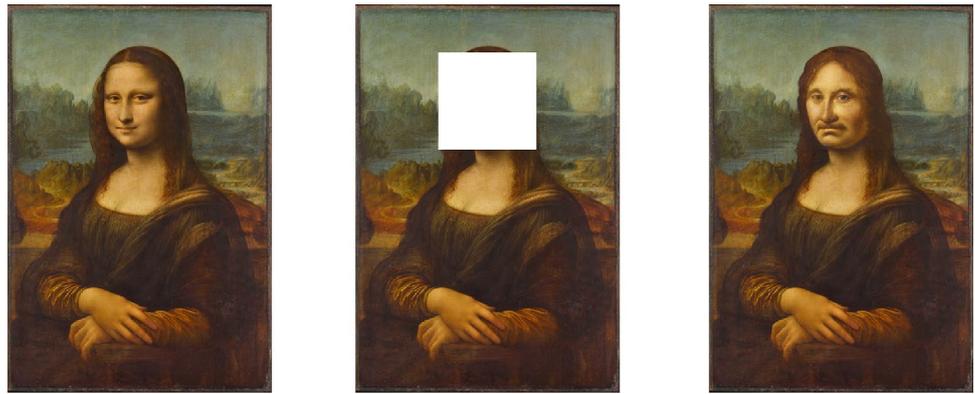

**Fig. 2** Example of the use of inpainting on "Leonardo Da Vinci, Mona Lisa, 1503", (left) original image, (center) image after applying a mask, (right) image after using inpainting to complete the missing fragment

In this work, we analyze the inference loop using a well-known AI image model, Stable Diffusion (Rombach et al. 2022a), and the inpainting functionality. The main contributions of the paper are:

1. To introduce the concept of recursive inpainting as an AI image transformation that can lead to information loss.
2. To propose a methodology to evaluate the impact of recursive inpainting using random sequences of masks applied to the images.
3. To conduct an extensive experimental evaluation of recursive inpainting with Stable Diffusion (Rombach et al. 2022a) using different mask sizes, number of iterations and images.
4. To analyze the results, discuss the limitations and propose several ideas to extend the study of recursive inpainting.

The rest of the paper is organized as follows: in Sect. 2, the inpainting functionality and the generative AI loops are briefly described. Then, the inference loop, denoted as Recursive Inpaiting (RIP), is presented in Sect. 3 and then evaluated in Sect. 4. The limitations of our evaluation, as well as the results, are discussed in Sect. 5. The paper ends with the conclusion in Sect. 6.

## 2 Preliminaries

### 2.1 Inpainting

One of the functionalities implemented by some modern generative AI image tools is inpainting (Yang et al. 2023), which takes an image with missing fragments and fills in those fragments to complete the image (Lugmayr et al. 2022; Jin et al. 2023). An example of the use of inpainting is illustrated in Fig. 2. In this case, Stable Diffusion was used, and we started from a complete image, applied a mask to remove some parts, and then used inpainting to complete the image. This enables a comparison between the original image and the result of inpainting. It can be seen that the tool is able to produce an image that resembles the original one. Interestingly, the AI model used, Stable Diffusion, changes the face to one that resembles a male which matches theories about the painting being a portrait of one of Leonardo Apprentices.[2] Different runs produce results with different types of faces, mostly woman-like.

The performance of inpainting depends on the model, the type of image, and the sizes and locations of the missing fragments (Moral-Andrés et al. 2023). In general, of the information lost in the image fragments, inpainting can only recover a fraction. A number of metrics can be used to measure the similarity between the original image and the reconstructed one (Quan et al. 2024): from classical ones such as the Structural Similarity (SSIM) (Wang et al. 2004) or the multi-scale SSIM (MS-SSIM) (Wang et al. 2003) based on the pixel level, to more advanced ones such as the Learned Perceptual Image Patch Similarity (LPIPS) (Zhang et al. 2018) or the Paired/Unpaired Inception Discriminative Score (P/U-IDS) (Zhao et al. 2021), which use AI models to capture human-like perceptual aspects.

### 2.2 Recursiveness in generative AI

The massive use of generative AI to generate text and images is creating a loop in which AI-generated content is uploaded to the Internet and then scrapped to train newer AI models (Martínez et al. 2023). This can lead to performance degradation of AI models or even to their collapse when they are trained with data produced by themselves (Dohmatob et al. 2024). This has triggered interest in understanding under which conditions these generative AI models are stable when trained recursively with data produced by the AI models (Bertrand et al. 2023; Briesch et al. 2023). This depends on several factors, including the model, the amount of AI-generated data used for each retraining, and whether the loop includes a single or several AI models. The study of this loop is important as it may impact both future AI models but also the nature of future content that will dominate the Internet. In all these studies, recursiveness involves training newer AI models with data generated from other AI models.

---

[2] https://www.cbsnews.com/news/male-model-behind-the-mona-lisa-expert-claims

The use of AI-generated data for training is not the only concern, since AI-generated data enter the Internet, it is likely to be processed again and again by other AI models, for example, to produce variations of it. This can lead to information loss or degeneration when AI models are not able to preserve the main elements of the data they operate with. For example, imagine an article that is scrapped from the Internet, then translated into another language using an AI model and posted on the Internet again. Subsequently, another AI model paraphrases the text and posts it again, and so on. Eventually, this can lead to information loss and even misinformation. This second scenario in which information is processed recursively poses a significant challenge that deserves to be studied in detail. However, the best of our knowledge, there are no previous works that have considered the impact of recursively processing images with AI models. This work is an initial step in the study of this problem.

## 3 Recursive inpainting (RIP)

An interesting observation is that a recursive loop for AI image models can be created when using inpainting. This is illustrated in Fig. 3; we start from an image and then apply a mask to remove some parts of it and use inpainting to complete them. At this point, we have a second image that the AI image model has partly created. Then, we repeat the process using a different mask to obtain a second image that, in this case, is created from AI-generated content. The process continues, and we recursively apply inpainting on images that have already been inpainted. In the process, as we remove and reconstruct parts of the images, information will be lost, but will this lead to images that are completely different from the original? images that are simpler and less complex? or will the inpainting be stable and lead to images that are only variations of the original image? This recursive inpainting process provides a simple scenario to study the potential implications of recursive information processing by AI models and can be used as an initial step in its study.

An example of recursive inpainting is shown in Fig. 4. The top left plot corresponds to the original image, in this case, a portrait of Pope Innocent X by Velázquez. The other images correspond to the results after applying inpainting two, four, six,..., up to sixteen times on one-fourth of the image. It can be seen that as the iterations progress, the image starts to depart from the original, and significant changes are introduced. However, even after the sixteen iterations, the final image still resembles the original one. Instead, when the same process is done for a sketch by Vincent van Gogh, as shown in Fig. 5 the lady in the original image no longer appears in the last image, which is completely different from the initial one.

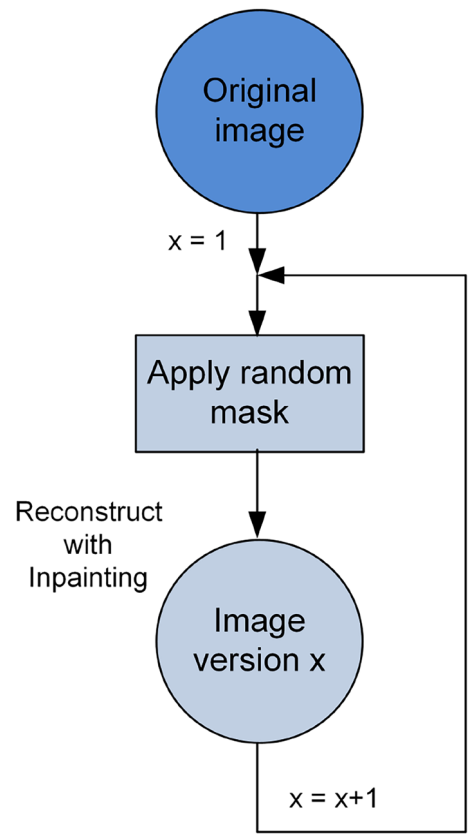

**Fig. 3** Illustration of the Recursive InPainting (RIP) process

The impact of recursive inpainting depends on many factors, such as the AI model, the type of image used, or the masks applied at each iteration. Intuitively, more complex images or masks that remove larger parts of the image will be more likely to lead to collapse. In the following section, the findings of an extensive empirical study of recursive inpainting with Stable Diffusion are presented as a first step toward understanding the key factors that determine the impact of recursive inpainting.

## 4 Evaluation

The main parameters for the recursive inpainting are:

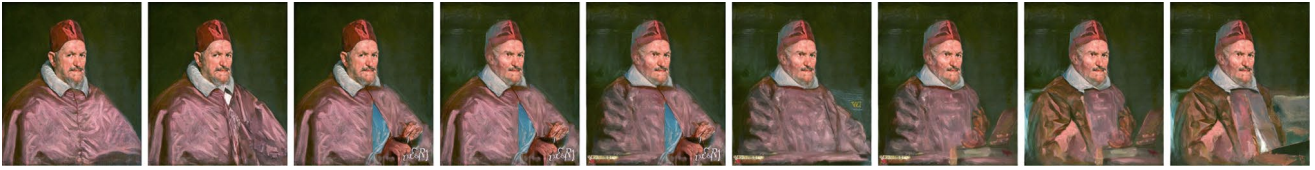

**Fig. 4** Example of recursive inpainting on "Diego Velázquez, Portrait of Innocence X, 1650". On the left, the original image is shown. Each subsequent image to the right displays the result after applying two recursive inpainting operations, up to the final image after sixteen inpainting operations

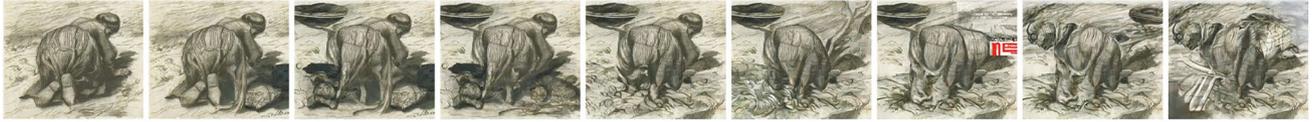

**Fig. 5** Example of recursive inpainting on "Vincent Van Gogh, Lugekone, 1885", On the left, the original image is shown. Each subsequent image to the right displays the result after applying two recursive inpainting operations, up to the final image after sixteen inpainting operations

1. The AI model.
2. The input images.
3. The masks applied at each step.
4. The number of iterations.

In our experiments, Stable Diffusion (Rombach et al. 2022a), a latent text-to-image diffusion model (Rombach et al. 2022b; Sohl-Dickstein et al. 2015), has been used because it is an open model and one of the most widely used AI image models. In particular, a version of Stable Diffusion 2 fine-tuned for inpainting was used.[3] This model employs a mask-generation technique (Suvorov et al. 2022) where the masked regions, along with the latent variational autoencoder (VAE) representations of the masked image, serve as additional conditioning for the inpainting process. The model parameters were set to the default values. Finally, to make the model focus on reconstructing the missing parts from the remaining visual elements, no text prompt was used to guide the painting.

As for the images, two sets are used in the evaluation:

1. A set of 100 randomly chosen images taken from a large dataset with more than 81,0000 art images of several types and made by different artists.[4] The goal of this set is to avoid bias in the selection of the images.
2. A set of 60 random real photos taken from the coco dataset.[5] This set will be used to have a reference with real photos since the aim of the manuscript is to compare the effect of inpainting in different art styles.
3. A set of 80 images taken from seven pictorial styles and from sketches of an architect in subsets to 10 images. This third set aims to evaluate how the recursive inpainting process affects different styles.

The first evaluation set and the results of the experiments for that set are available, both image and metrics as the source images are taken from a public dataset. For the second set, only the results in terms of metrics are public as copyrights protect some images. Additionally, the description of the images is provided so that the experiments can be reproduced. The code used to run all experiments including all the settings used in Stable Diffusion is also available in the public repository.[6] The input images are 512 × 512 pixels. when their original form factor is not square, blank bands are added on the sides to fit the 512 × 512 pixels format.

To generate the masks for inpainting, the images are divided into squares of a given size, and in each iteration, a square is randomly selected and used as the mask. The generation of the masks is illustrated in Fig. 6 for the case of a 128 × 128 square and two iterations. It can be observed that one square is removed in each iteration. Then, inpainting is run, and the results obtained for the pixels in the mask are used to replace the ones in the initial picture. This modified picture is then used as the input image for the next iteration. This procedure guarantees that at each iteration, the inpainting only modifies the pixels in the selected mask.

---

[3] https://huggingface.co/stabilityai/stable-diffusion-2-inpainting
[4] https://huggingface.co/datasets/huggan/wikiart
[5] https://cocodataset.org/
[6] https://doi.org/10.5281/zenodo.11532111

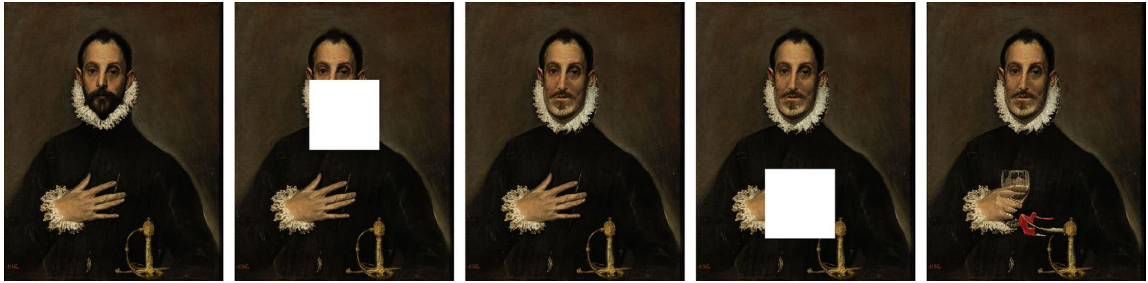

**Fig. 6** Example of recursive inpainting on "El Greco, The Nobleman with his Hand on his Chest, 1580"

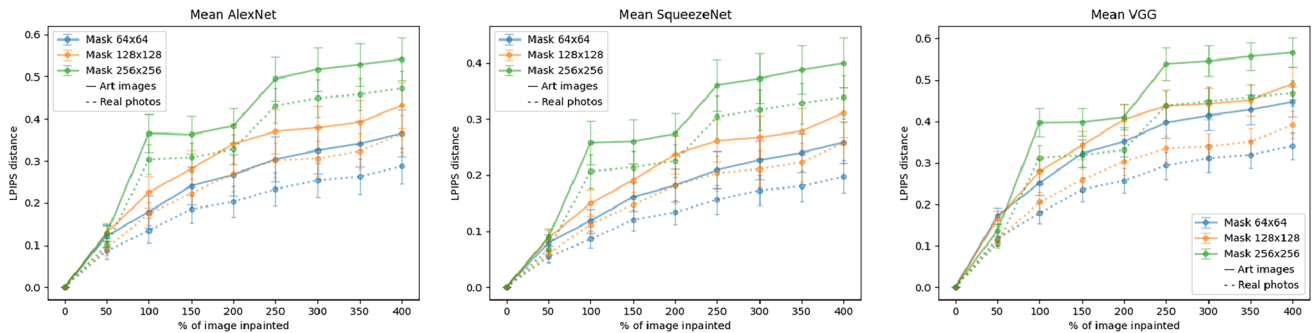

**Fig. 7** Mean LPIPS across the 100 art images and 60 real photos versus the inpainting for three neural networks: AlexNet (left), SqueezeNet (middle) and VGG (right) for different mask sizes (64 × 64, 128 × 128, 256 × 256) with 95% confidence interval bars

To estimate the similarity with the original image across iterations, we use the Learned Perceptual Image Path Similarity (LPIPS) (Zhang et al. 2018) metric widely used to assess the quality of inpainting.[7] In the implementation used, the features of three neural networks can be used to compute the metric: SqueezeNet (Iandola et al. 2016), AlexNet (Krizhevsky et al. 2012), and VGG (Yu et al. 2016).

To enable a direct comparison of inpainting with different mask sizes, our experiments use as the main parameter not the number of inpainting operations but the number of pixels on which inpainting is done. For example, for a 256 × 256 mask, four inpainting operations correspond to changing a number of pixels equal to those in the original 512 × 512 images. Instead, for a 128 × 128 mask, sixteen inpainting operations are needed to change 512 × 512 pixels.[8] Using as a metric the number of pixels that are inpainted relative to the image size makes comparisons easier across different masks and image sizes.

### 4.1 Random images

The first part of the evaluation focuses on the set of randomly selected images to try to extract conclusions on the impact of recursive inpainting that applies to images in general. In the first experiment, we take the 100 random art images and the 60 random photos and perform recursive inpainting for 400% of the pixels with masks of 64 × 64, 128 × 128, 256 × 256. To quantify the degradation as inpainting operations are done, the LPIPS metric between the original image and each generation has been computed using the three neural networks (SqueezeNet, AlexNet, and VGG) features. The goal is to get an initial understanding of the degradation trends and also of the neural network to use for computing the LPIPS metric in the rest of the experiments. The average distances in the 100 art images and 60 real photos at each step of 50% inpainting are shown in Fig. 7. The bars show the 95% confidence interval computed on the samples on each of the data points and are intended to illustrate the variability of the values. Several initial observations can be made from the results:

1. As recursive inpainting progresses, the distance with the original image increases. This could eventually lead to an image that bears no resemblance to the original.

---
[7] The implementation used is available in a public repository https://github.com/richzhang/PerceptualSimilarity

[8] Note that the pixels changed in two iterations can be the same as each iteration selects a mask randomly.

2. The slope of the distance tends to become smaller but does not seem to stabilize even when the distance is large.
3. The difference with the original image is larger when the size of the mask used for inpainting is larger which as discussed before is in line with the intuition that it is harder to inpaint larger blocks.
4. The three networks used to compute the LPIPS (SqueezeNet, AlexNet, and VGG), provide similar results.
5. The confidence intervals are large which suggests that different behaviors will be observed for different images.
6. The behavior is the same in both art images and real photos, but real photos do not diverge as much as art images.

To better understand the variability of the distances for each image, scatter plots of the LPIPS distances of the 100 (60) art images (real photos) for each of the neural networks are shown in Fig. 8. It can be observed that there is significant variability across images but the trends are similar to the ones observed in the mean: distance is larger with more inpainting and with larger masks. Comparing the three networks (SqueezeNet, AlexNet, and VGG), the last one, VGG is the one with fewer outliers. VGG is also the most complex network and thus should be expected to better capture the

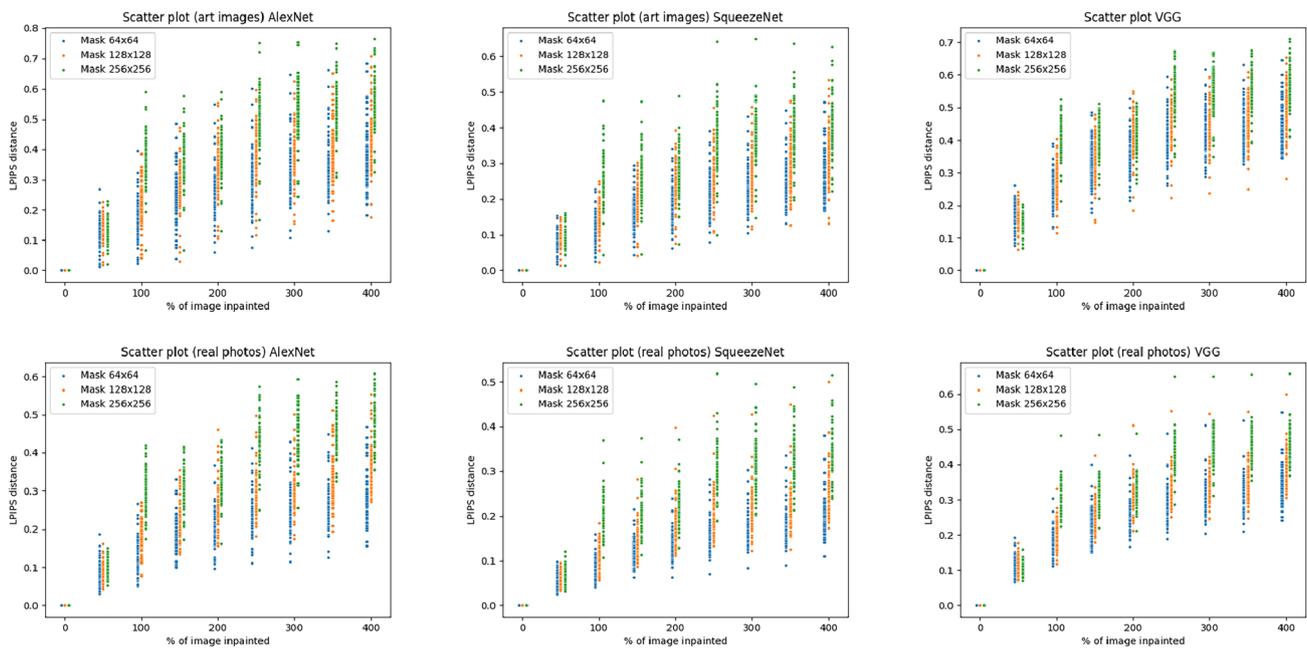

**Fig. 8** Scatter plots of the LPIPS across the 100 art images and 60 real photos versus the inpainting for three neural networks: AlexNet (left), SqueezeNet (middle) and VGG (right) for different mask sizes (64 × 64, 128 × 128, 256 × 256)

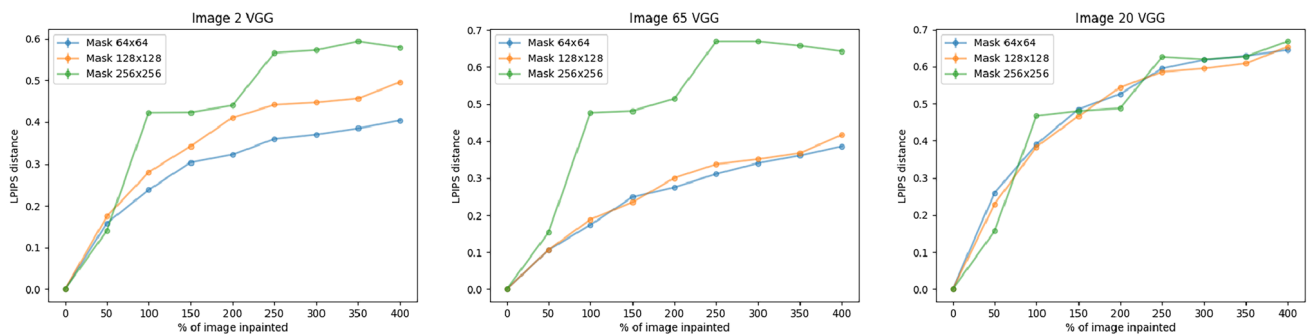

**Fig. 9** Examples of different types of behaviors of the LPIPS metrics during the recursive inpainting: Distance stabilizes (left), Large distance and variation with mask size (middle), similar distance with mask size (right)

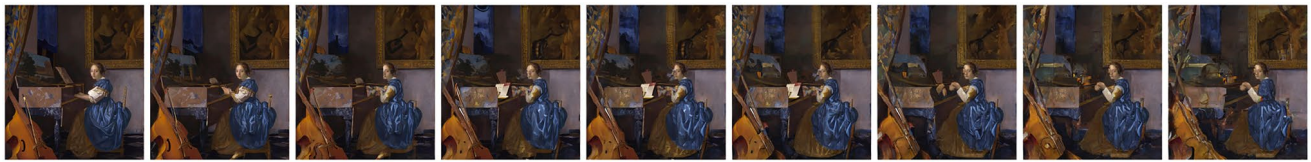

(a) 64x64

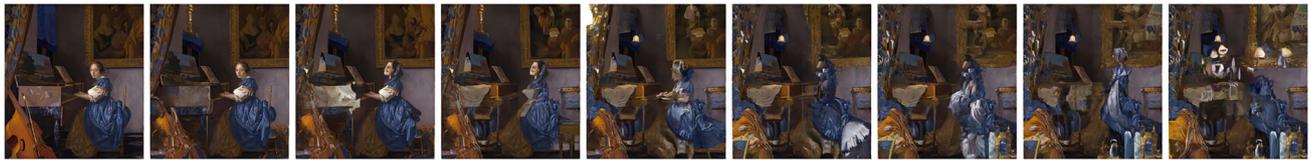

(b) 128x128

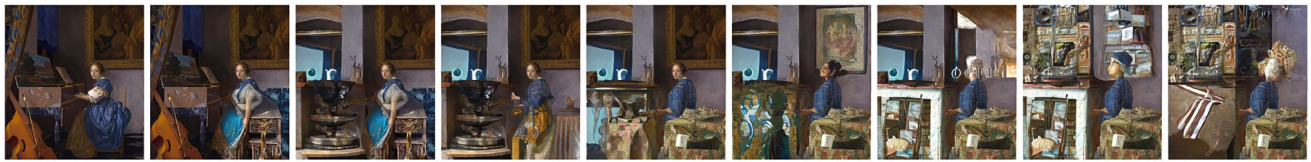

(c) 256x256

**Fig. 10** Results of the recursive inpainting with different mask sizes for the image corresponding to Figure 9 (left), "Johannes Vermeer van Delft, Lady Seated at a Virginal, c. 1672"

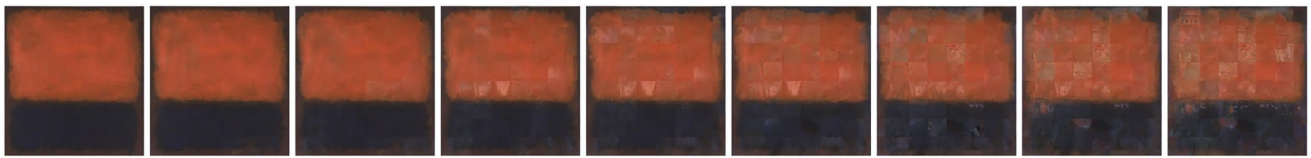

(a) 64x64

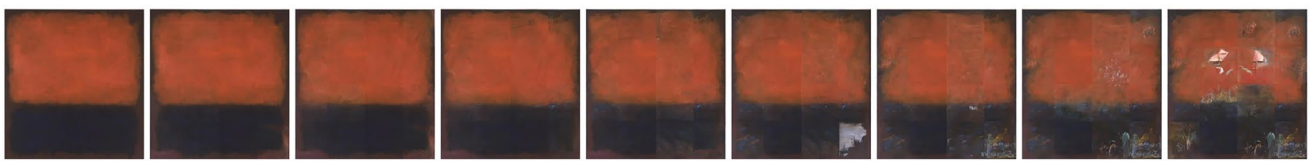

(b) 128x128

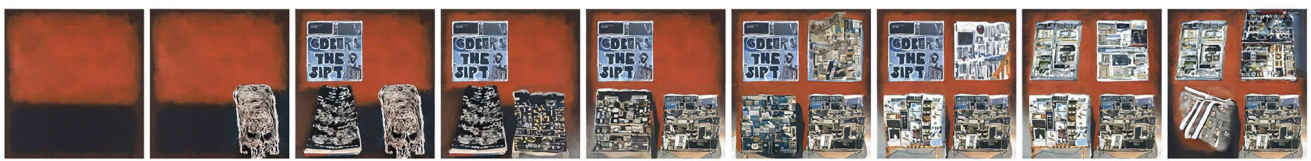

(c) 256x256

**Fig. 11** Results of the recursive inpainting with different mask sizes for the image corresponding to Figure 9 (middle), "Mark Rothko No. 14, 1960"

features of the images (Yu et al. 2016). Therefore, in the following, we only report results for VGG although all the metrics are available in the repository along with the images.

Another factor that can impact the degradation is the image used as the starting point for the process. To analyze this, LPIPS distance plots were generated for each image and analyzed manually. A few illustrative examples are shown in Fig. 9. In the first one (left), the distance tends to stabilize as recursive inpainting progresses. In the second, there is a large difference in the distances with mask size and finally, in the last one, the distances are similar for all mask sizes. The image sequences for the three images are shown in Figs. 10, 11 and 12.

In the first image even when using a small mask, see Fig. 10a

For the second image, the results in Fig. 11 show the impact of the mask size, when it is small, Fig. 11a

The second experiment explores the variability of the impact of recursive inpainting when applied to the same image. As the parts removed are randomly chosen, it is of interest to see whether the degradation is similar on different runs. To understand the variability of the degradation with the run, 10 images have been selected from the set of 100, and each has been run 10 times. The LPIPS metrics across runs for three different images are shown in Fig. 13 when using the VGG network which again tends to have the lowest deviations. It can be observed that the variations are larger for larger masks which is expected as the larger the mask, the fewer the iterations to reach a given percentage of inpainting which causes more variability. The variations are also reduced as the percentage of inpainting increases showing again, that the larger the number of inpainting operations the lower the variability.

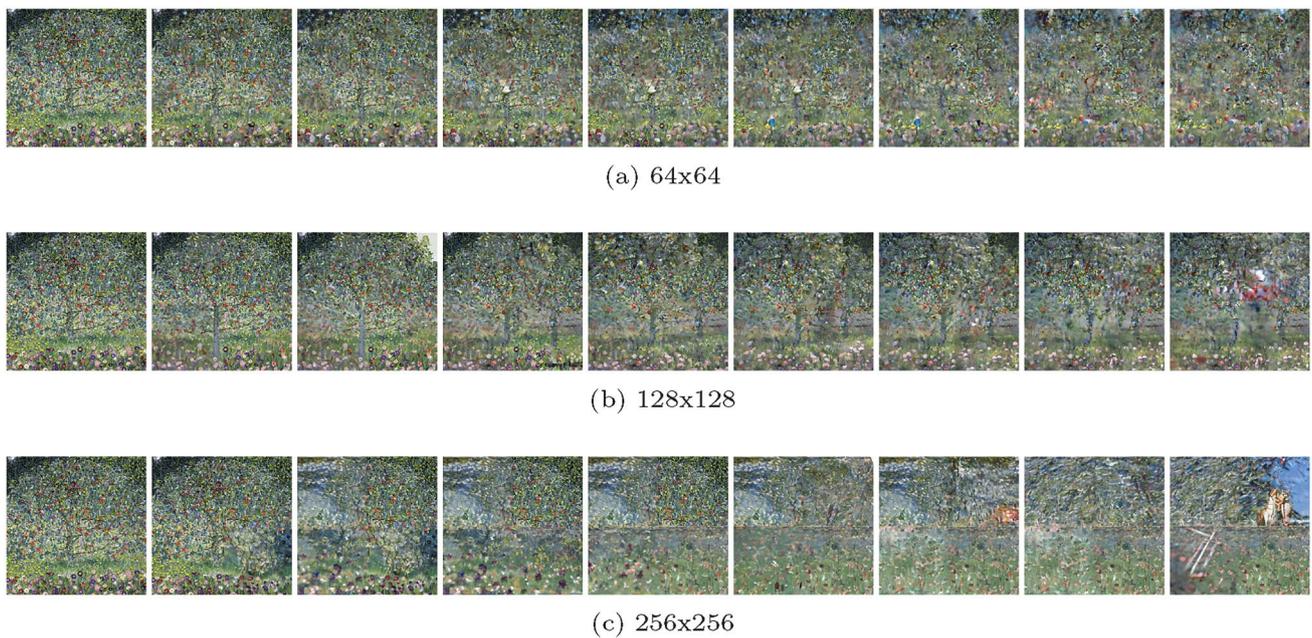

(a) 64x64

(b) 128x128

(c) 256x256

**Fig. 12** Results of the recursive inpainting with different mask sizes for the image corresponding to Figure 9 (right), "Gustav Klimt, The Apple Tree, 1912"

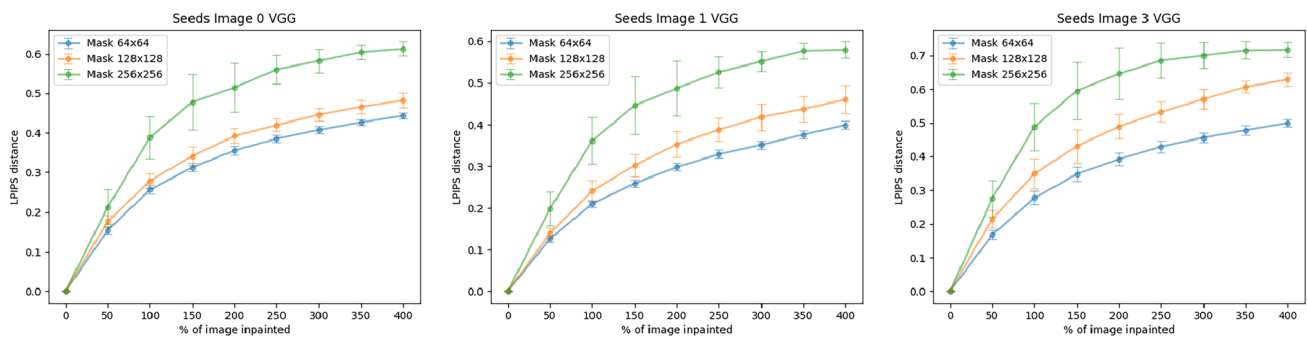

**Fig. 13** LPIPS for ten runs with different seeds on the same image for three different images and mask sizes ($64 \times 64$, $128 \times 128$, $256 \times 256$) with 95% confidence interval bars

This means that recursive inpainting seems to converge in terms of LPIPS distance as the process advances.

Finally, looking at the results in qualitative terms from an aesthetic perspective, the results raise several important concerns, independent of the quantitative analysis of the metrics. It is surprising when the AI, instead of fixing the missing parts, creates new things that don't fit in the painting. This likely happens when small bits left over from the erased sections make the AI assume an object was there, even if it wasn't. This shows the AI doesn't truly recognize the painting it is dealing, with which raises doubts about the entire result. Moreover, the AI seems to not know the rules for making things look realistic in terms of perspective. It twists angles in odd ways at times. The worst cases happen when the AI cannot comprehend how faces are rotated and tries to reconstruct them haphazardly, just to make it resemble a head or a human component. In some images, the AI attempts to reconstruct a sort of color palette in the best situations, particularly when the alterations are minimal. However, in other cases, it simply uses arbitrary colors or elements to try to resemble the original image which results in a pixelated appearance.

### 4.2 Pictorial styles

The third set of images is divided into eight different styles, seven pictorial and the last one architectural each having 10 images.[9] The selection of images aims to create a varied group of elements with a specific structure from a stylistic point of view. This allows for a collection of elements that, while not necessarily from the same author, share a creative context that facilitates comparative analysis. The diversity of styles force the inpainting to understand and reconstruct different shapes, coloring and objects. The objective is to understand whether recursive inpainting has a different impact depending on the style of the images. Again the dataset of 60 photos has been used to compare the art images with real images. The art styles included in the set are:

- Pre-Raphaelitism: presents a selection characteristic of this movement: paintings with a pictorial sensitivity focused on parameters predating those of Raphael Sanzio. The abundance of human figures, with faces showing restrained expressions, is set within lush vegetation. The colors are vivid, and the defining details are abundant in all the clothing and objects depicted in these images. The relationship established between the background, typically rural, and the figures is significant, as it allows most paintings to be read with only two levels of depth.
- Expressionism: characterized by strongly contrasting colors. Striking at every level, at times violent, supported by clear, geometrically defined lines, these works depict very particular scenes. Human figures are stripped of their subtler nuances, with their most pronounced features emphasized. They create unsettling atmospheres, and in some cases, feel deliberately unfinished, as if waiting to reach a more defined stage.
- Historicism: features significant groups of people in various key moments of the nation's history. These paintings clearly narrate notable events in an iconic manner, with characters depicted with distinctive faces and unique expressions, far removed from stoic realism, to reveal the magnitude of the events portrayed. The characters are the protagonists, taking precedence over the locations where they are set.There is a dual interpretation to consider that defines this collection: how digital tools operate when dealing with extensively group-oriented scenes versus others with a smaller number of figures.
- Realism: consists of 20 th-century Spanish painters within the orbit of their main representative, Antonio López. All of these are urban images, whether in close-up or from broader perspectives, and none feature human figures. However, there are buildings, vegetation, and even everyday objects. The definition of the outlines is not always equally precise. This fact is considered interesting because, within the same stylistic body, having variants linked to the pictorial technique adds greater complexity to the analysis of the work.
- Cubism: encompasses a small selection where both people and objects are broken down into more or less recognizable geometric forms. These geometries also construct the various perceptual planes of the executed images. In this predominantly object-focused group, a limited number of pieces have been selected where human elements are also treated with similar codes. However, the attributes are clearly distinguishable, whether dealing with living beings or inanimate objects.
- Neoplasticism: the most radical body in terms of formal construction with orthogonal geometric structures and a limited color palette, all executed with very precise definition in terms of lines, angles, and the boundaries of colored surfaces. It is also important to note the significance of white, which, in most of the selection, serves as a large neutral base on which the rest of the black and colored elements of the compositions are built.
- Pop Art: presents significant variations in form within the style's typical values. Everyday objects appear through collages, texts, comic-like structures, and a reduced color palette, all participating in different representative works of this trend. A key aspect is the varied formalization of the human figure, with prominent close-ups, along with the simplification of elements that appear in other pieces.

---

[9] For this second set, the description of the images is provided in the public repository.

- Architectural sketches: this final set is characterized by black line drawings representing all types of new infrastructures for the contemporary city. This collection is of particular interest as it depicts buildings in perspective, but without color, except for the tone of the paper, which has changed over time. It also involves considering structures that repeat different patterns, as seen, for example, in the top of buildings, which are key to understanding the specific characteristics of the author's style.

The entirety of the works has aimed to combine a proper reading at the level of constructing a specific stylistic corpus while presenting distinct and limited singularities that might trigger a challenge for recursive inpainting. In more detail, the images cover different aspects:

- The selection has used images that have not been modified with any type of filter.
- The selection has been based on works of notable relevance in the history of each style.
- The selection seeks to create defined and measurable fields of study, avoiding any data dispersion.
- The selection is considered appropriate for generating an analytical set organized around each of the different styles presented.
- The selection has sought to include recurring motifs, such as the human figure, nature, or geometry, but approached by the artists in different ways.
- The selection has aimed to unite images with varying levels of detail and different framing on urban or figurative themes.
- The selection has aimed to work with both basic and complex geometries, as well as with precise and more diffuse definitions.
- The selection aims to avoid archetypal and recurrent images that may be easier to reconstruct.

As for the first set of random images we take the 80 images from the different styles and 60 photos from the coco dataset and perform recursive inpainting for 400% of the pixels with masks of $64 \times 64$, $128 \times 128$, $256 \times 256$ and the LPIPS metric between the original image and each generation is used to measure the degradation. The average distances on the 10 images for each of the pictorial styles at each step of 50% inpainting are shown in Fig. 14. As in previous experiments, the bars show the 95% confidence interval computed on the samples on each of the data points and are intended to illustrate the variability of the values.

The following observations can be made from the results:

1. The degradation increases with the percentage of inpainting as for random images.
2. The degradation is larger for larger mask sizes as for random images.
3. The styles that are closer to realistic images such as Realism or Historicism suffer less degradation than the rest.
4. The less realistic styles, such as Expressionism or Pop Art suffer more degradation.
5. Qualitatively, the trends are similar for all styles. Real photos are close to realism and historicism, which are the styles that represent reality most faithfully. These styles also tend to show less deviation.

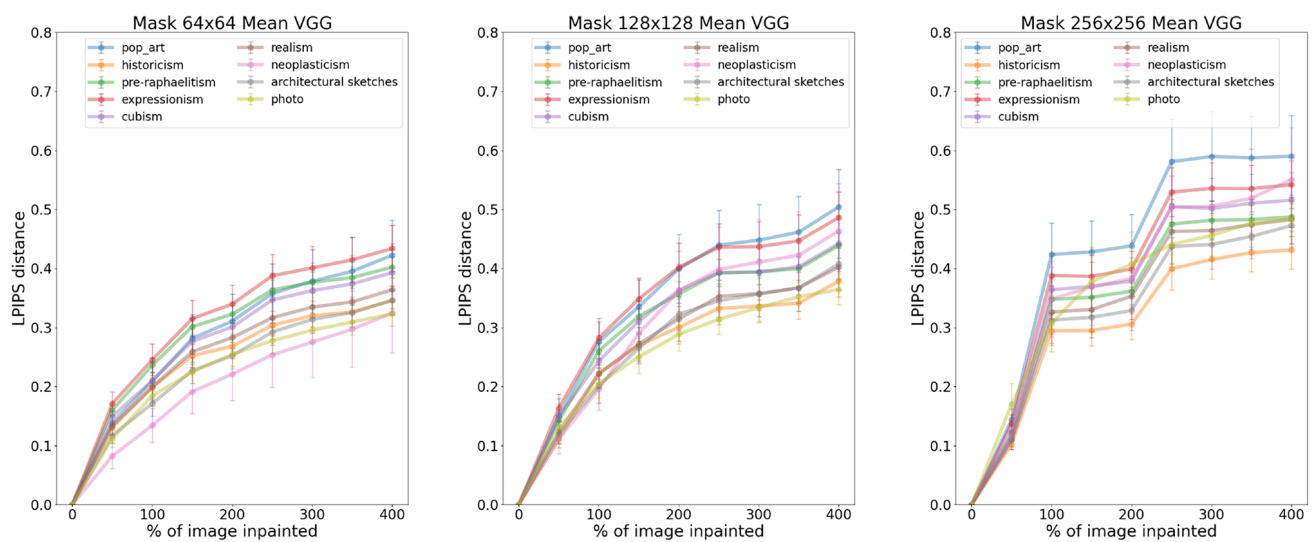

**Fig. 14** Mean LPIPS across the 10 images for each pictorial style and 60 real photos versus the inpainting done for VGG for different mask sizes ($64 \times 64$, $128 \times 128$, $256 \times 256$) with 95% confidence interval bars

In addition to the use of the LPIPS distance, the images have been visually analyzed by an expert and the following qualitative observations can be made:

1. The images in many cases remain relatively stable up to number three or number six in the results grid. Those correspond to inpainting of 100% and 250% of the number of pixels in the image. Beyond that point, the presented deformations are notable. This is consistent with a large increase in the LPIPS distance observed for the 256x256 mask size from 200% to 250%. It can be seen that only when the inpainting retains the majority of the initial traits the image is similar to the original. When this is not the case, the results are more original and deviate from the model.
2. Deformations and reconstructions can be differentiated, more or less successfully, that maintain the spirit of the style, as opposed to others that are proposed as aberrations due to the lack of connection that can be made with the initial object on which the work was based.
3. Generally speaking, the deformations found in the gestures of human figures and in their faces are notable, at times recreating structures completely alien to the original painting. This is significant since figures have not always been represented in the same way. However, the results tend to be more peculiar and deviate more from the initial model compared to those obtained with objects.
4. The reconstructions carried out on original paintings with few details, simple geometries, or sparse coloration lead to reconstructions that are closer to the original element. This seemingly evident fact needs to be tested precisely to confirm the process compared to others that affect more complex structures.
5. Those paintings that feature straight and curved lines lead to reconstructions of two levels: more accurate in the straight realm, and more aberrant in the curved realm. This connects to the case of people, where different geometries naturally combine and are less segregated.
6. In the reconstruction system, purely creative results are also produced by the algorithm used, generating formal realities that are far removed from the reference works employed. This opens up creative, rather than reconstructive, interpretations.

Therefore, it seems there is always a degradation but it tends to be smaller for realistic images. A possible reason is that inpainting models have been trained primarily on real images, so they work better with this kind of art style.

In this regard, we have taken the real images from the COCO dataset and categorized them. The COCO dataset is used to train object detection models, so it contains labels of objects that appear in the images. We have filtered the images based on the objects they contain in an exclusive manner, resulting in six categories, each consisting of 10 images: apples, elephants, fire-hydrants, toiltes, persons, trains.

Fig. 15 shows the VGG LPIPS distances of the different categories in different mask sizes. After a visual analysis of the images and through distance metrics, the following results were observed:

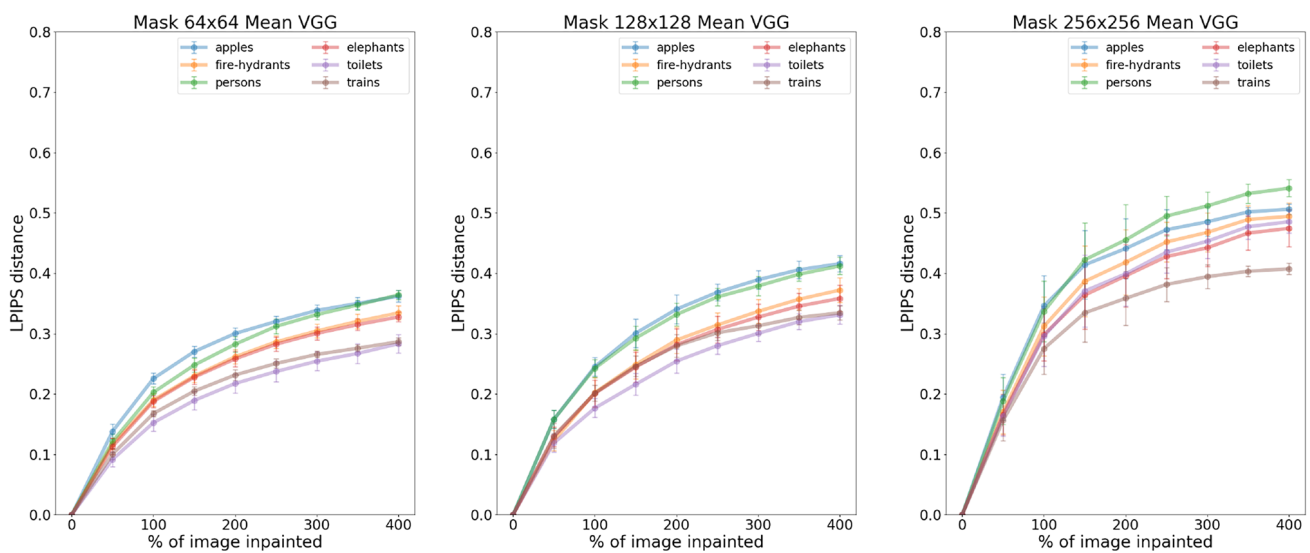

**Fig. 15** Mean LPIPS across the 10 images for each category of real images versus the inpainting done for VGG for different mask sizes (64 × 64, 128 × 128, 256 × 256) with 95% confidence interval bars

1. All categories follow a similar pattern to that of art images.
2. People and apples are the categories that remain the least stable. Complex entities such as people and small objects like apples appear to be more sensitive to inpainting. The former because their physical features, posture, etc., can be altered, which is reinforced through iterations. In the case of apples, these are small objects where the mask can cover the entire object, causing the reference to be lost and replaced by other objects or deleted.
3. Big and robust elements like elephants and trains are the most stable categories. One possible reason is that they are structures with a simple geometry, symmetrical, homogeneous, and without many details. Additionally, in these images, trains and elephants tend to occupy a large portion of the image, making inpainting more robust by using the rest of the image as a reference as it is shown in Fig. 16.
4. Medium-sized and homogeneous objects like toilets remain stable with small masks but not with large ones. This is because, with large mask sizes, the entire toilet is replaced, whereas with small mask sizes, the toilet

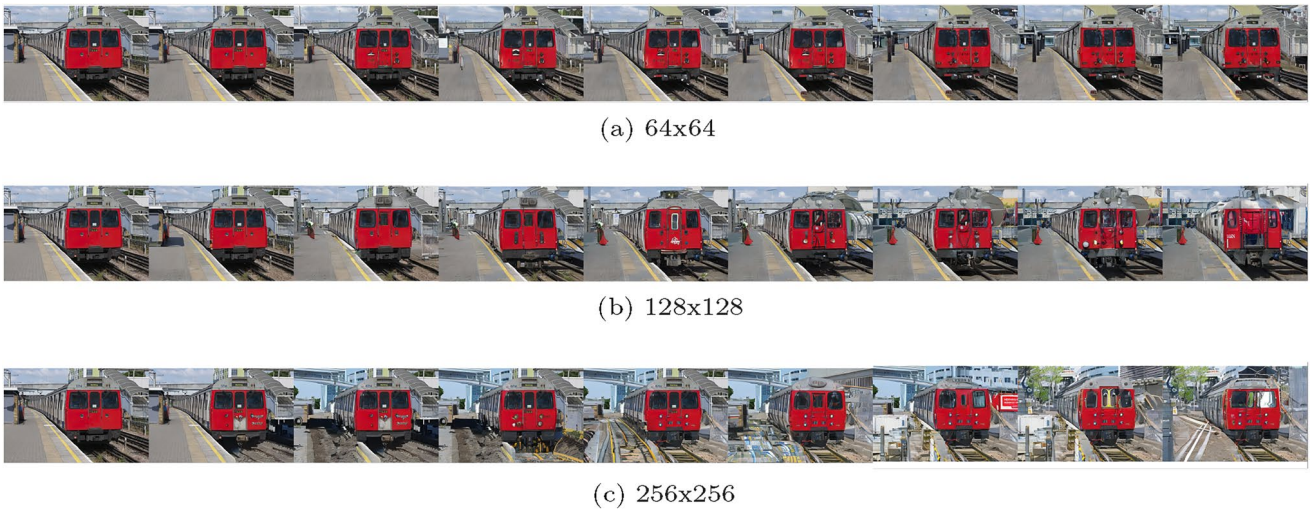

(a) 64x64

(b) 128x128

(c) 256x256

**Fig. 16** Results of the recursive inpainting with different mask sizes for the image corresponding to Figure 503985 in coco dataset (photo of a train)

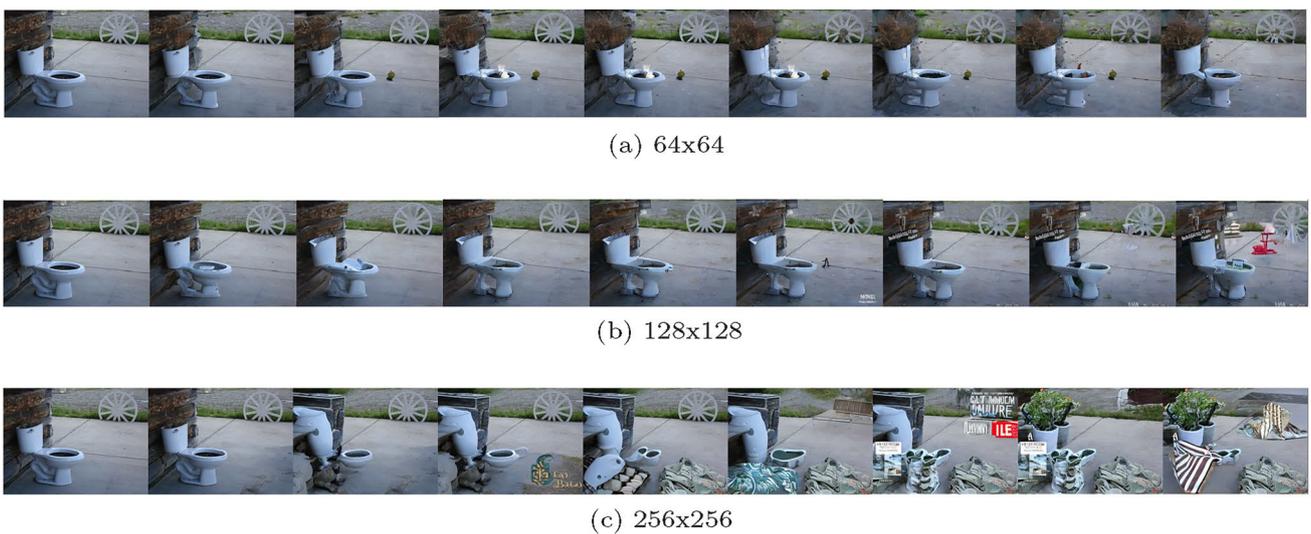

(a) 64x64

(b) 128x128

(c) 256x256

**Fig. 17** Results of the recursive inpainting with different mask sizes for the image corresponding to Figure 224010 in coco dataset (photo of a toilet)

is only partially reconstructed, leaving little chance for variability. An example of this effect is shown in Fig. 17.

### 4.3 Impact of colors

In the last part of our evaluation, we explore whether colors affect the degradation caused by recursive inpainting. To that end, from each original image, three images are generated by removing one of the three components of color, Red, Green or Blue (RGB) and a fourth image by removing all colors so that the image is on black and white. Then, the recursive inpainting experiments with the same settings as in the previous section is run.

Three different patterns are observed when comparing the impact of RIP on the original image and the color modifications. The first pattern occurs for Realism, Cubism, Pre-raphaelitism, Expressionism, and Historicism and is illustrated in Fig. 18 that shows the results for Historicism. It can be seen that for small mask sizes there are only small differences in the distances between the original and the modified images. Instead for a mask of size 256x256 the distances are larger when removing a color component while black-and-white images have slightly lower distances than the originals.

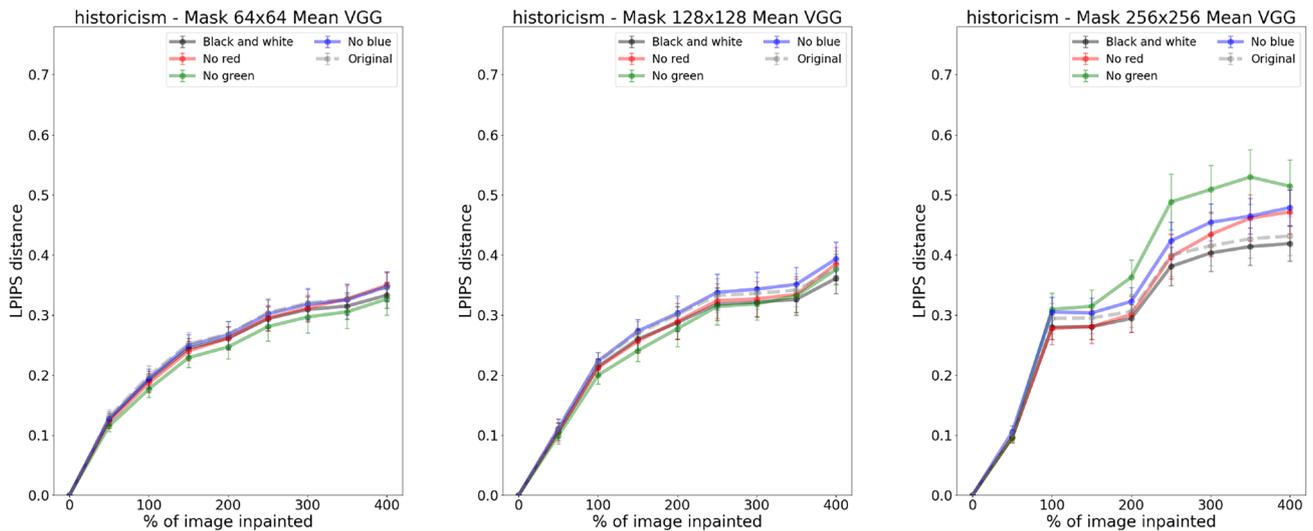

**Fig. 18** Mean LPIPS across the 10 Historicism images for original, black and white, and color modifications versus the inpainting done for VGG for different mask sizes (64 × 64, 128 × 128, 256 × 256) with 95% confidence interval bars

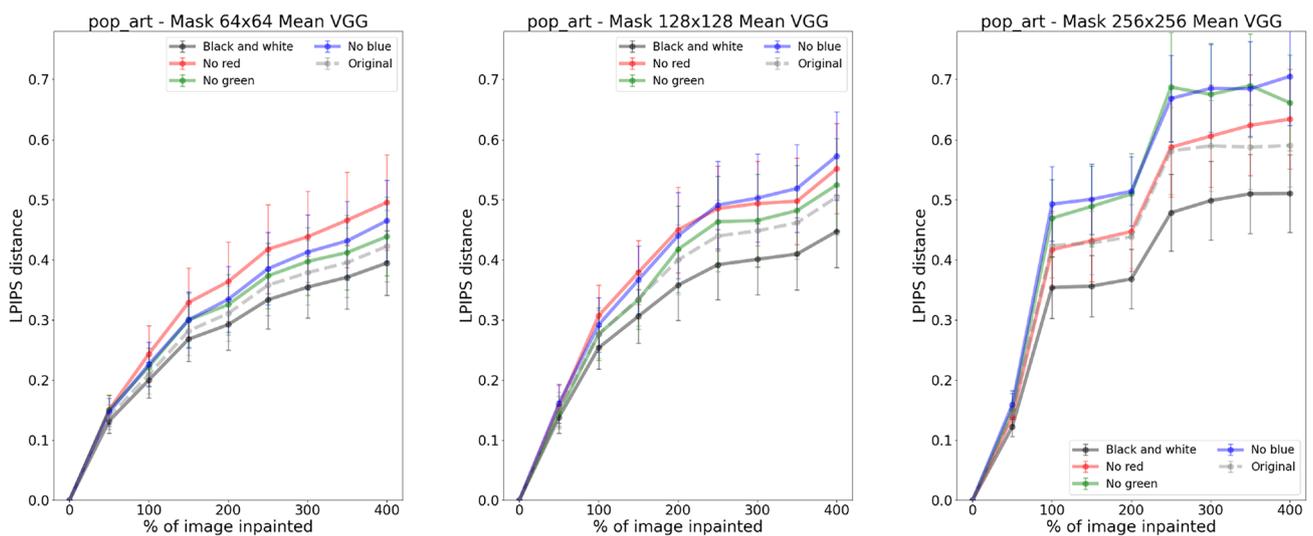

**Fig. 19** Mean LPIPS across the 10 Pop Art images for original, black and white, and color modifications versus the inpainting done for VGG for different mask sizes (64 × 64, 128 × 128, 256 × 256) with 95% confidence interval bars

The second pattern occurs for Neoplasticism and Pop Art and is illustrated for pop art in Fig. 19. In this case, the impact of removing colors is significant for all mask sizes and the black and white images have significantly lower distances than than the original ones.

Finally, the Architectural Sketches show a different pattern from the rest as shown in Fig. 20. This may be due to the fact that the original images are almost colorless. It can be observed that for Neoplasticism and Pop Art, the impact of removing colors is significant for all mask sizes but differently from those styles, black and white images have the largest distances for small mask sizes.

From the results of the different styles, it seems that removing color components tends to have a lower impact for smaller mask sizes and depends significantly on the image type. For images that are already colorless or close, the impact seems to be different as observed in the case of architectural sketches with black and white images having a larger impact.

## 5 Limitations and discussion

The evaluation carried out is just an initial step in understanding the effects of recursive painting and has several limitations:

1. The study is entirely empirical with no theoretical analysis of recursive inpainting.
2. The results presented are only for one AI model, Stable Diffusion. Other diffusion models or other model types such as Generative Adversarial Networks (Creswell et al. 2018) should be evaluated to understand if the results presented are specific to Stable Diffusion or are more generally applicable. Additionally, for each model the impact of the hyper-parameters on recursive painting should also be studied.
3. The procedure used to implement recursive inpainting applies random masks of the same size at each iteration. It would be interesting to evaluate alternative schemes, for example using different mask sizes at different iterations or using deterministic sequences.
4. The selection of images used in the evaluation corresponds mostly to artworks of different artists and styles and some photos. This may introduce bias in the inpainting process. A larger number of images, possibly with different features in terms of the objects represented and their shapes and sizes, should be evaluated to have a better understanding of how the impact of recursive inpainting varies with image type.
5. The LPIPS is used to compare images, it would be interesting to use additional metrics in the comparison such as the ones discussed in Sect. 2.

In addition to extending the evaluation and developing a theoretical analysis for recursive inpainting, it is of interest to design mitigation strategies to try to reduce information loss. From our experiments, a general observation is that smaller mask sizes tend to reduce image degradation compared to larger mask sizes when applied to the same number of pixels. Therefore, smaller mask sizes could be used for mitigation when that is possible in the application.

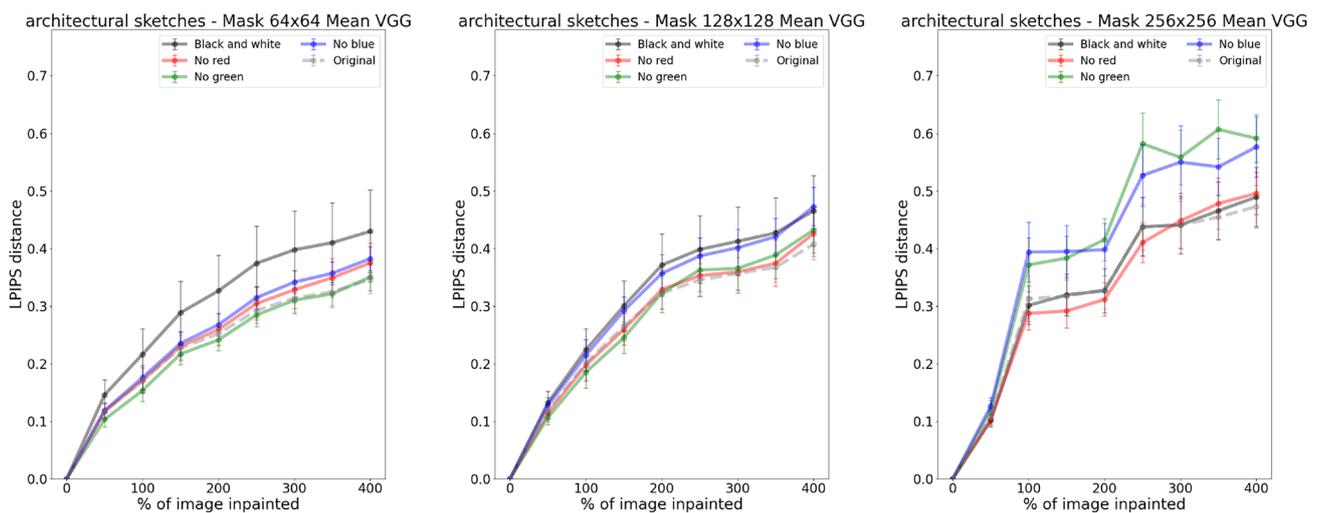

**Fig. 20** Mean LPIPS across the 10 Architectural Sketches images for original, black and white, and color modifications versus the inpainting done for VGG for different mask sizes (64 × 64, 128 × 128, 256 × 256) with 95% confidence interval bars

Another approach when the process is done by the same tool or organization could be to compare the original and final images and, when there is significant degradation, automatically discard the image or at least flag the degradation to the user. The study of mitigation strategies is an interesting area for future work.

Another area of study would be to compare the results of AI with those produced by humans when faced with the same problem. However, doing experiments with humans would require a significant effort and also depend on their painting capabilities. To apply recursive inpainting, several persons, one per iteration would be needed to make sure that they have not seen the original or previous images in the series which makes the procedure rather complicated. In summary, as discussed before, this paper is just the first step in the analysis of recursive inpainting that is primarily intended to present the problem and motivate further work.

Even with the limitations discussed, the results presented show how recursive inpainting can lead to images that are completely different from the original ones. This illustrates the potential of AI to transform content and lead, even if unintentionally, to information loss. This is similar to the model collapse observed when training generative AI models with their own data (Marchi et al. 2024) for which techniques to avoid collapse are being proposed (Gerstgrasser et al. 2024). Analyzing the similarities and differences between recursive inpainting and recursive training loops is another avenue for future research. Exploring modifications to the AI models to mitigate information loss in recursive inpainting is also of interest and could lead to better image-generation AI models. More broadly, understanding if there is a relationship between the model collapse effect on recursive training and the degeneration of the image in the recursive inpainting is also of interest.

Finally, although recursive inpainting has been proposed as a tool to study the potential information loss of AI inpainting models, its use to produce variations or transformations of an image may also be an interesting area for future work that to the best of our knowledge has not been explored so far.

## 6 Conclusion and future work

In this paper, the effect of recursive inpainting on AI image models has been presented and empirically studied as an example of the potential effects of successive transformation of content by AI tools. The results for Stable Diffusion show that recursiveness can lead to images that are significantly different from the original and even to the degeneration of the image into something not related to the original. Although further experiments are needed to evaluate other models, images, procedures, and configurations, this is similar to what has been observed in the recursive training of generative AI models, which is attracting significant interest from the community. Therefore, this paper opens another area in the research of the impact of the recursive use of generative AI, in this case only in the inference phase and focusing on the information loss that AI tools may introduce as they recursively process content.

The analysis of recursive inpainting presented in this paper is just the first step. Additional AI models, images, and model configurations should be tested to better understand the impacts of recursive inpainting. Beyond empirical results, it is also of interest to develop theoretical models that can explain the impacts of recursive inpainting. Exploring the links between recursive training and recursive inpainting is also an interesting area for future research.

**Acknowledgements** This work was supported by the FUN4DATE (PID2022-136684OB-C21/22), ENTRUDIT (TED2021-130118B-I00), and SMARTY (PCI2024-153434) projects funded by the Spanish Agencia Estatal de Investigación (AEI), and by the Chips Act Joint Undertaking project SMARTY (Grant no. 101140087).

**Funding** Open Access funding provided thanks to the CRUE-CSIC agreement with Springer Nature.

**Data availability** The data (or links to the data sources) used in this paper, as well as the code to reproduce the experiments, are available in a public repository at https://zenodo.org/records/15098223.